\documentclass[letterpaper, 10 pt, conference]{ieeeconf}  

\IEEEoverridecommandlockouts                              

\usepackage{amsmath} 
\usepackage{graphicx}
\usepackage{color}
\usepackage{caption}
\usepackage{subcaption}
\usepackage{float}
\usepackage[absolute,overlay]{textpos}
\title{\LARGE \bf
Grabbing power line conductors based on the measurements of the magnetic field strength
}

\author{Goran Vasiljevi{\'{c}}$^{1}$, Dean Martinovi{\'{c}}$^{1}$, Matko Orsag$^{1}$ and Stjepan Bogdan$^{1}$
\thanks{ This work was supported by the project  AERIAL COgnitive Integrated Multi-task Robotic System with Extended Operation Range and Safety (AERIAL CORE) EU-H2020-ICT (grant agreement No. 871479)}
\thanks{$^{1}$Goran Vasiljevi{\'{c}}, Dean Martinovi{\'{c}}, Matko Orsag and Stjepan Bogdan are with University of Zagreb
Faculty of Electrical Engineering and Computing, Laboratory for Robotics and Intelligent Control Systems (LARICS), Unska 3, Zagreb 10000, Croatia; 
        {\tt\small goran.vasiljevic@fer.hr}}
}

\begin{document}

\maketitle
\thispagestyle{empty}
\pagestyle{empty}
\maxdeadcycles=20000
\begin{textblock*}{14.9cm}(3.2cm,0.75cm) %
	{\footnotesize © 2021 IEEE.  Personal use of this material is permitted.  Permission from IEEE must be obtained for all other uses, in any current or future media, including reprinting/republishing this material for advertising or promotional purposes, creating new collective works, for resale or redistribution to servers or lists, or reuse of any copyrighted component of this work in other works.}
\end{textblock*}

\begin{abstract}

This paper presents the method for the localization and grabbing of the long straight conductor based only on the magnetic field generated by the alternating current flowing through the conductor. The method uses two magnetometers mounted on the robot arm end-effector for localization. This location is then used to determine needed robot movement in order to grab the conductor. The method was tested in the laboratory conditions using the Schunk LWA 4P 6-axis robot arm.

\end{abstract}

\section{INTRODUCTION}
\setcounter{footnote}{1}
The Aerial Core Project\footnote{https://aerial-core.eu/} is aimed at the maintenance and monitoring of large infrastructure. It includes, among other things, the maintenance of high-voltage lines using aerial manipulation. Examples of this are the setting of spacers for high-voltage lines and helical or clip-type bird diverters. Some of these actions require high forces to be generated by the manipulator, and for this purpose additional manipulator would be added to the system, where one manipulator would be used to hold the power line conductor while the other manipulator performs the required manipulation operation. These operations can generally be carried out while the power line is in operation, generating a strong alternating magnetic field around the conductor. This could cause interference in the electronics and sensors of the aerial manipulator. The idea of this paper is to show how it is possible to use this magnetic field as an alternative localization method of high-voltage line. Ultimatively we demonstrate the approach to detect and grab the conductor of the power line relying solely on the measurements of the magnetic field.

Enabling aerial robots to interact with the environment is not a novel concept, but rather a fast developing field of research \cite{ollero2018aeroarms}. In this work we follow the best practices in this field \cite{korpela2014hardware}, to emulate the design through an off-the-shelf standalone manipulator, before moving to aerial manipulation. There are several systems that rely on the magnetic field for localization and navigation. Most of the work is designed for indoor environments with special infrastructure for magnetic navigation. Such an approach is shown in \cite{brown1973} and \cite{Everett1995} for vehicle navigation, which follows a wire that conducts alternating current laid under the floor based on two components of the magnetic field. This type of navigation is available in commercial autonomous guided vehicles. A similar approach is presented in \cite{Kamewaka1987}, where the magnetic guidance band is placed on the floor for localization, which is detected by a magnetic flux sensor.

In \cite{martinovic_electric_2014} a positioning system for inductive charging of electric vehicles was developed. For this purpose, at least two magnetometers are mounted on the underbody of the vehicle to sense the low-frequency magnetic field emitted by the charging coil in the parking lot. The vehicle then locates the charging coil by applying the trilateration principle to the measured data \cite{martinovic_magnetic_2015}. Since the magnetization effects in the ferromagnetic underbody can be compensated \cite{martinovic_electric_2014-1},\cite{martinovic_dealing_2019}, the assistance system achieves millimeter accuracy.

Several state of the art localization solutions rely on magnetic beacons distributed in the environment for 2D and 3D localization, such as the systems presented in \cite{Sheinker2013}, \cite{Sheinker20132}, \cite{Sheinker2016} and \cite{Mitterer2018}. A similar approach is presented in \cite{Son2016} for 5D localization, where the robot is equipped with magnets and the Hall effect sensors are distributed in the environment.
\begin{figure}[t]
	\centering
	\includegraphics[width=0.995\linewidth]{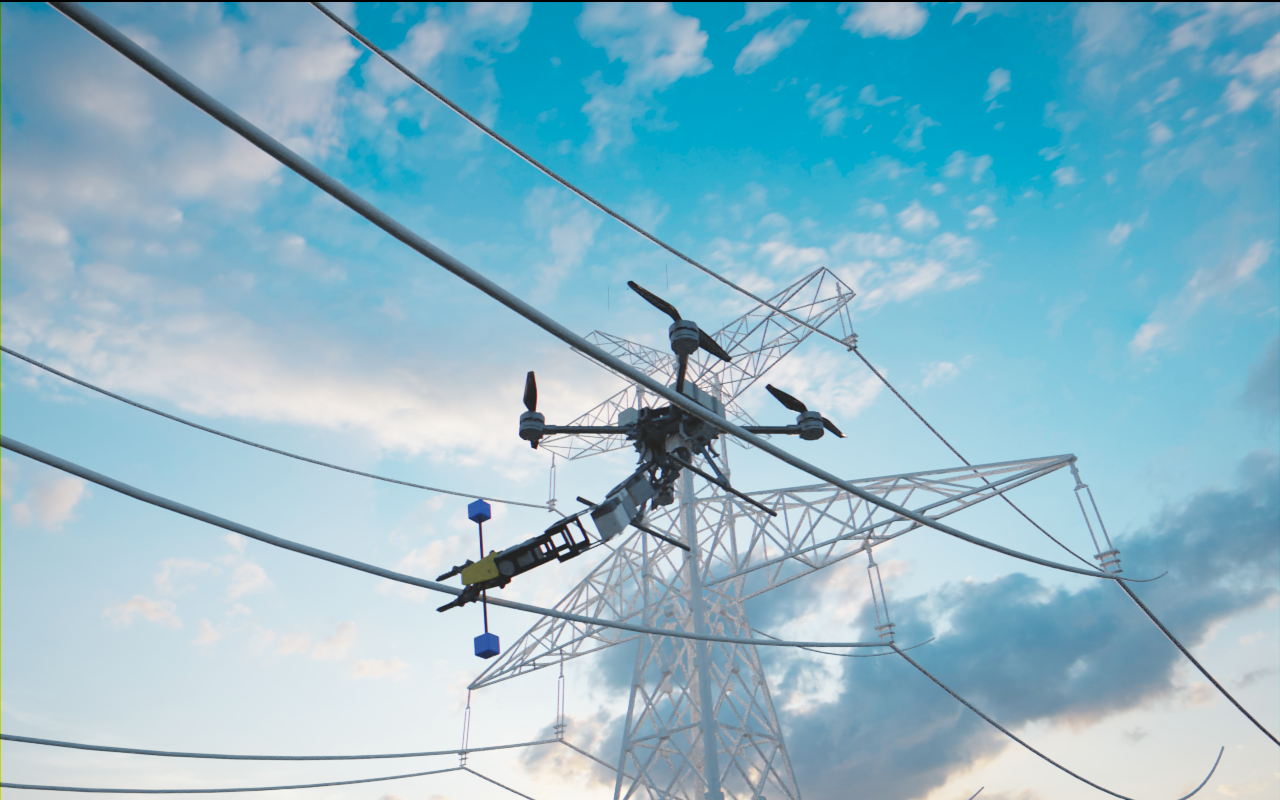}
	 \caption{AERIAL-CORE project focuses on developing an integrated aerial cognitive robotic system for applications such as the inspection and maintenance of large powerline infrastructures.}
     \label{fig:aerial_core}
\vspace{-0.5cm}
\end{figure}
A special type of localization based on magnetic fields is used in medicine to track robotic endoscopic capsules. Such applications are shown in \cite{Than2012}, \cite{Song2014}, \cite{Popek2017}, \cite{Hu2005}, \cite{Hu2007} and \cite{hu2010}. The number of works deals with localization based on the geomagnetic field maps of the environment. Examples of such work are \cite{Vallivaara2011}, \cite{Wang2016}, \cite{Lee2018}, \cite{Akai2017}, \cite{Frassl2013} and \cite{Hanley2017}.

In \cite{martinovic_mathematical_2021} it is shown that unmanned aerial vehicles (UAVs) can precisely navigate through the magnetic field of two long parallel power lines using at least three magnetometers. The core result is the analytic solution of the corresponding nonlinear magnetic field equation. As a precondition the current in the lines must be known.

The conductor of a power line can be considered as a long, rigid, cylindrical object. There are a number of papers dealing with the gripping of known but also unknown objects. The authors in \cite{Richtsfeld2008} present a method for recognising and gripping cylindrical objects.
In \cite{Sintov2019} the authors present a method for  regrasping elongated objects. In \cite{Koo2011} a method for a wire harness assembly motion planner is presented, and the authors introduce a method for gripping a wire. A similar problem is considered in \cite{Qin2019}, where authors use humanoid robots for cable installation and describe, among other things, a method for gripping a deformable cable. In \cite{Zapata-Impata2019}, a method for calculating gripping points using three-dimensional point clouds is presented, which is demonstrated by gripping the cylindrical object. In \cite{Seo2017}, authors present a gripping of a cylindrical object using visual servoing by an aerial manipulator.

In this paper we present a method for locating and grabbing the straight conductor carrying alternating current, which is based only on the magnetic field measurement of two 3-axis magnetometers mounted on the 6-axis robot end-effector. To the authors' knowledge, such a system has never been implemented before.

The paper is organized as follows. In the following section we describe method for localization of conductor based on the magnetic field strength using two 3-axis magnetometer. Section III. presents a method for grabbing the line in 3D space.  Section V. presents experimental results in laboratory environment. Finally, Section VI. gives the short conclusion of the work.

\section{MAGNETIC FIELD BASED LOCALIZATION OF POWER-LINE CONDUCTOR}
\label{sec:localization}

This section describes the localization of the power line conductor based on the magnetic field generated by the current flowing through the conductor. The current flowing through the power line is an alternating current of sinusoidal form with a frequency of 50 Hz or 60 Hz, which means that the magnetic field around the conductor is also sinusoidal.

\subsection{Magnetic field around the conductor}

In the vacuum, the magnetic field strength generated by the current flowing through the straight long conductor is:
\begin{equation}
    B=\frac{\mu_0I}{2\pi r}
    \label{eq::magneticfield}
\end{equation}
where $B$ is the magnetic field strength, $I$ is the current flowing through the conductor, $r$ is the distance from the conductor and $\mu_0$ is the permeability of the vacuum:
\begin{equation}
    \mu_0=4\pi 10^{-7}H/m
\end{equation}
The equation \eqref{eq::magneticfield} gives the strength of the magnetic field at a distance $r$ from the conductor, and its direction vector is based on the the right hand rule with respect to the conductor (see Figure \ref{fig:aerial_grabbing}). The vector will always have a direction that is circumferential in relation to the isomagnetic lines.

If we define that the conductor is a line in 3D, it can be described with the following equation
\begin{equation}
    \mathbf{x_{pl}}=\mathbf{x_{pl0}}+\mathbf{v_{pl}}t
    \label{eq::line}
\end{equation}
where $\mathbf{x_{pl}}=\begin{bmatrix} x_{pl} &  y_{pl} & z_{pl}\end{bmatrix}^T$ defines points on a  line, $\mathbf{x_{pl0}}=\begin{bmatrix}x_{pl0} & y_{pl0} & z_{pl0}\end{bmatrix}^T$ is one point on the  line, $\mathbf{v_{pl}}=\begin{bmatrix}v_{xpl0} & v_{ypl0} & v_{zpl0}\end{bmatrix}^T$ is a line direction vector and $t$ is a scalar.

The magnetic field vector at a point $\mathbf{x_0}=\begin{bmatrix}x_0 & y_0 & z_0\end{bmatrix}^T$ in space can be defined:
\begin{equation}
    \mathbf{B}=\frac{\mu_0I}{2\pi \vert\mathbf{x_0}-\mathbf{x_{pl1}}\vert}\frac{(\mathbf{x_0}-\mathbf{x_{pl1}}) \times \mathbf{v_{pl}}}{\vert(\mathbf{x_0}-\mathbf{x_{pl1}}) \times \mathbf{v_{pl}}\vert}
    \label{eq::field_vector}
\end{equation}
where $\mathbf{x_{pl1}}$ is the point on the conductor closest to the point $\mathbf{x_0}$, which can be calculated using the equation \eqref{eq::line} with $t$ calculated in the following way:
\begin{equation}
    t=-\frac{(\mathbf{x_{pl0}}-\mathbf{x_0})\cdot \mathbf{v_{pl}}}{\vert \mathbf{v_{pl}}\vert^2}
\end{equation}

\subsection{Localization based on two 3-axis magnetometer}

The position of the power line can be determined with two 3-axis magnetometers at known relative poses:
\begin{equation}
    \mathbf{T_{m_1}^{m_2}}
    =\begin{bmatrix}
        \mathbf{R} & \mathbf{p}\\
        \mathbf{0} & 1
    \end{bmatrix}
    \label{eq::magnetometertransformation}
\end{equation}
where $T_{m_1}^{m_2}$ is homogeneous transformation matrix , $\mathbf{R}$ is a rotation matrix and $\mathbf{p}$ is a translation vector from the coordinate system of magnetometer $m_1$ to the coordinate system of magnetometer $m_2$.

The measurements of the magnetometer are magnetic field strength vectors resulting from \eqref{eq::field_vector}  in 3 axes:
\begin{equation}
\begin{array}{ll}
    \mathbf{B_{m_1measured}}=
    \begin{bmatrix}
        b_{x1} & b_{y1} & b_{z1}
    \end{bmatrix}^T\\
    \mathbf{B_{m_2measured}}=
    \begin{bmatrix}
        b_{x2} & b_{y2} &  b_{z2}
    \end{bmatrix}^T
    \end{array}
\end{equation}

To use the equation \eqref{eq::field_vector}, the two vectors must be expressed in the same coordinate system. For this purpose, we can take the coordinate system of the first magnetometer as a common coordinate system ($\mathbf{B_{m_1}}=\mathbf{B_{m_1measured}}$), so that only the measurements of magnetometer 2 need to be transformed, which is done by

\begin{equation}
    \mathbf{B_{m_2}}=\mathbf{R}^{-1}\mathbf{B_{m_2measured}}
\end{equation}
From the equation \eqref{eq::field_vector} it can be seen that the two vectors $\mathbf{B_{m1}}$ and $\mathbf{B_{m2}}$ are perpendicular to the vector $\mathbf{v_{pl}}$, because of cross product with $\mathbf{v_{pl}}$ in the equation. Assuming that the vectors $\mathbf{B_{m1}}$ and $\mathbf{B_{m2}}$ are not parallel, the conductor direction vector can be calculated.

For non-parallel magnetic field vectors, the conductor direction vector $\mathbf{v_{pl}}$ is calculated as follows:

\begin{equation}
    \mathbf{v_{pl}}=\frac{\mathbf{B_{m1}} \times \mathbf{B_{m2}}}{\vert\mathbf{B_{m1}} \times \mathbf{B_{m2}}\vert}
\end{equation}

Again from equation \eqref{eq::field_vector}, the vector pointing from each magnetometer towards the individual closest point on the power line is perpendicular to the magnetometer measurement vector and the line vector:
\begin{equation}
    \begin{array}{ll}
    \mathbf{v_{m_1pl}}=\frac{\mathbf{x_{m1}}-\mathbf{x_{plm1}}}{\vert \mathbf{x_{m2}}-\mathbf{x_{plm1}} \vert}
    =\frac{\mathbf{v_{pl}} \times \mathbf{B_{m_1}}}{\vert \mathbf{v_{pl}} \times \mathbf{B_{m_1}} \vert}\\
    \mathbf{v_{m_2pl}}=\frac{\mathbf{x_{m2}}-\mathbf{x_{plm2}}}{\vert \mathbf{x_{m2}}-\mathbf{x_{plm2}} \vert}
    =\frac{\mathbf{v_{pl}} \times \mathbf{B_{m_2}}}{\vert \mathbf{v_{pl}} \times \mathbf{B_{m_2}} \vert}\\
    \end{array}
    \label{eq::towards_conductor}
\end{equation}

where $\mathbf{x_{m1}}$ is a null vector, $\mathbf{x_{m2}}$ is $\mathbf{p}$ defined in the equation \eqref{eq::magnetometertransformation}, $\mathbf{x_{plm1}}$ is a point on the conductor closest to the point $\mathbf{x_{m1}}$ and $\mathbf{x_{plm2}}$ is a point on the conductor closest to the point $\mathbf{x_{m2}}$.

The problem of finding the points $\mathbf{x_{plm1}}$ and $\mathbf{x_{plm2}}$ comes down to the well-known problem of finding the nearest point between two lines, where each line is defined by the points $\mathbf{x_{m1}}$ and $\mathbf{x_{m2}}$ together with its respective vector pointing in the direction of the conductor. One of the solutions to this problem is:

\begin{equation}
    \begin{array}{ll}
        \mathbf{d}=(\mathbf{x_{m2}}-\mathbf{x_{m1}})\\
        \mathbf{r}=\mathbf{d} - (\mathbf{d} \cdot \mathbf{v_{m_1pl}})\mathbf{v_{m_1pl}}  + (\mathbf{d} \cdot \mathbf{v_{pl}}) \mathbf{v_{pl}} \\
        t=-\frac{\vert\mathbf{r}\vert}{\mathbf{v_{m_2pl}}\cdot \frac{\mathbf{r}}{\vert\mathbf{r}\vert}}\\
        \mathbf{x_{plm2}}=\mathbf{x_{m2}}+\mathbf{v_{m_2pl}}t
    \end{array}
\end{equation}
The power line conductor can be uniquely defined by the point $\mathbf{x_{plm2}}$ and the vector $\mathbf{v_{pl}}$. It can be noted that the calculation did not use the value of the magnetic field strength, but only the magnetic field vector in each point. The field strength value could be used for results verification.

\begin{figure}[th]
	\centering
	 \begin{subfigure}[b]{0.46\linewidth}
	\includegraphics[width=\linewidth]{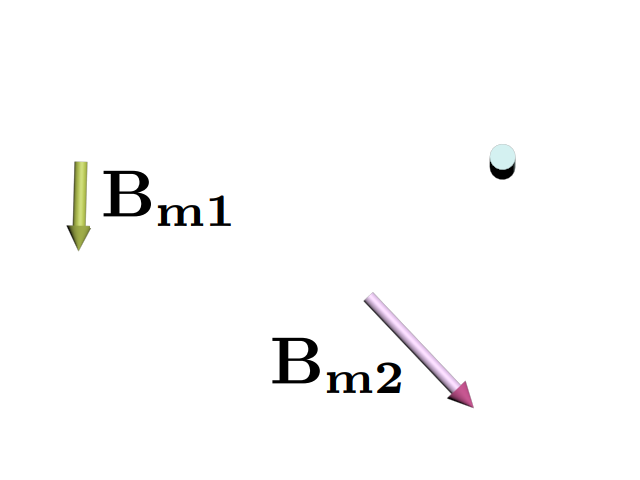}
	\caption{Top view  of  measured magnetic field vectors}
	 \end{subfigure}
	 \hfill
	 \begin{subfigure}[b]{0.46\linewidth}
	\includegraphics[width=\linewidth]{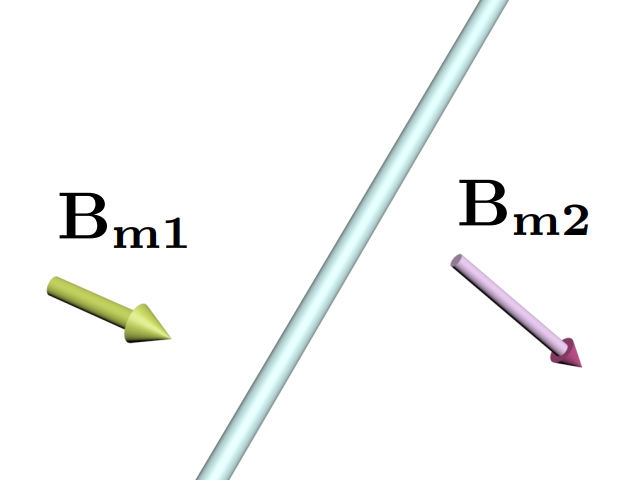}
	\caption{Free view of measured magnetic field vectors}
	 \end{subfigure}
	 \begin{subfigure}[b]{0.46\linewidth}
	\includegraphics[width=\linewidth]{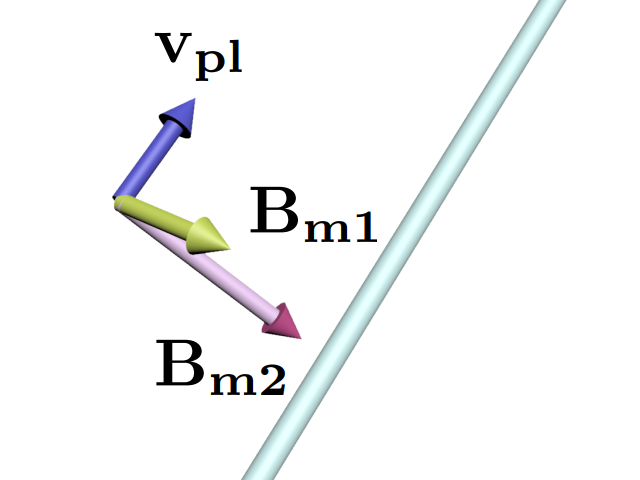}
	\caption{Cross product of vectors determines direction of the line}
	 \end{subfigure}
	 \hfill
	 \begin{subfigure}[b]{0.46\linewidth}
	\includegraphics[width=\linewidth]{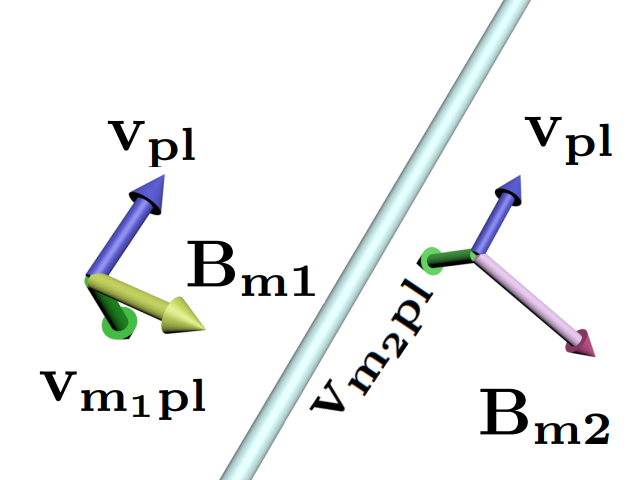}
	\caption{Calculating vectors pointing from magnetometers to  the conductor}
	 \end{subfigure}
	 \begin{subfigure}[b]{0.46\linewidth}
	\includegraphics[width=\linewidth]{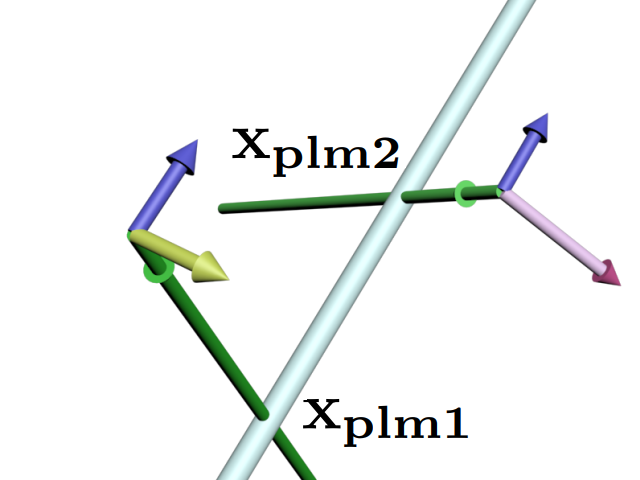}
	\caption{Calculating smallest distance between two lines - free view}
	 \end{subfigure}
	 \hfill
	 \begin{subfigure}[b]{0.46\linewidth}
	\includegraphics[width=\linewidth]{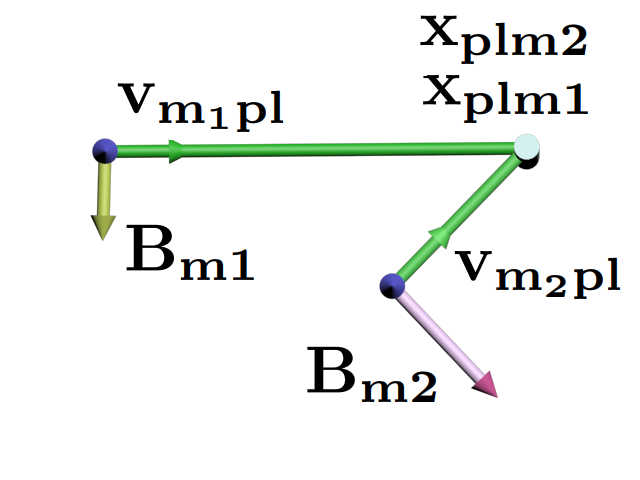}
	\caption{Calculating smallest distance between two lines - top view}
	 \end{subfigure}

	 \caption{Procedure of calculating conductor line based on magnetic field around conductor}
     \label{fig::localization_procedure}
\end{figure}

\subsection{Processing magnetometer readings in alternating magnetic field}

An alternating current flowing through the conductor generates the alternating magnetic field with the same frequency as the current. The magnetometer measures the  value of the magnetic field generated by the conductor  superimposed on the earth's magnetic field. For this reason, the magnetometer readings must be processed before the magnetic field vector can be used. Several different methods are known to extract components with defined frequency from the signal. One easy method is the Fourier transform (i.e. the discrete Fourier transform), which gives the signal strength of each frequency component. This method is well suited for extracting the signal strength of the specific frequency.

Another important aspect of the AC magnetic signal is that we cannot determine the direction of the signal, i.e. whether the signal is positive or negative. This is not really important, since the equations shown in the previous subsection are designed to work with positive and negative vector direction, but it is important that the relative direction relation between the vector components $x$, $y$, and $z$ is correct. The relative direction of vector components can be determined by checking whether the signals are in phase or antiphase. We always assume that the $x$ component of the vector is positive, then we check whether the $y$ and $z$ components are in phase or antiphase and conclude whether they have a negative or positive direction. This can be done by calculating phase offset obtained with the Fourier transform. If the difference between phase offset of the $x$ component and the $y$ component is greater than $\pi/2$ then the $y$ component has the opposite direction to the $x$ component of the vector. The same can be applied to the $x$ and $z$ components.  


\section{GRABBING CONDUCTOR}

The idea of the paper is to use the gripper mounted on the 6-axis robot arm to grab the conductor defined as a line in 3D. The grabbing of the line in 3D can be done in different ways depending on the grabbed point on the line and the approach angle of the tool. We present a method for selecting a point to be grabbed and the approach angle of the tool. In this paper we focus on grabbing the conductor with a robot arm with 6 degrees of freedom.

\subsection{Selection of the grabbing point}

The point on the line selected to be grabbed by the robot arm is the one closest to its base point $\mathbf{P_B}$.  

For a given line \eqref{eq::line}, the point $\mathbf{P_G}$ closest to $\mathbf{P_B}$ can be calculated:

\begin{equation}
    \begin{array}{ll}
        \mathbf{P_G}=\mathbf{x_{pl0}}+\mathbf{v_{pl}}t_g\\
        t_g=(\mathbf{x_{pl0}}-\mathbf{P_B})\cdot \mathbf{v_{pl}}
    \end{array}
    \label{eq::grabbingpoint}
\end{equation}

Using the equation \eqref{eq::grabbingpoint}, we can determine the point on the line closest to the robot base (in this case $\mathbf{P_B}$ is a zero vector), which can then be used to select the grabbing pose.

\begin{figure}[th]
	\centering
	\includegraphics[width=0.97\linewidth]{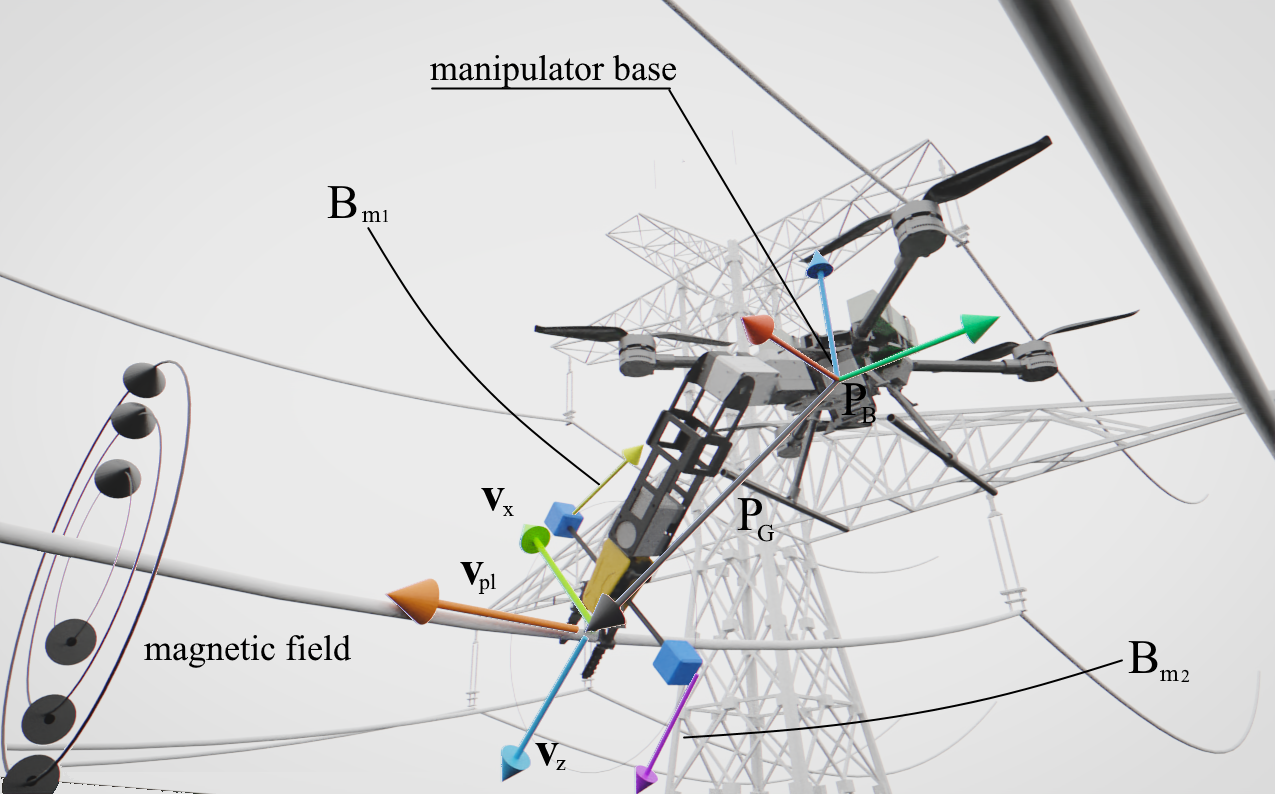}
	 \caption{Grabbing conductor with robotic arm}
     \label{fig:aerial_grabbing}
\end{figure}

\subsection{Grabbing orientation}

Given the grabbing position $\mathbf{P_G}$ and the line vector $\mathbf{v_{pl}}$, the pose of the end effector (grabbing pose) is not uniquely defined, but it depends on the approach angle, so we have to choose a grabbing orientation.

As already mentioned, the conductor direction is defined with the vector $\mathbf{v_{pl}}$. We choose to have the $y$ orientation of the end-effector point to the conductor vector $\mathbf{v_{pl}}$. To fully define an orientation, we need to choose another vector that is perpendicular to the conductor. The choice of this vector depends on the relative position of the grabbing position to the robot base.

One of the solutions to this problem is to select the vector that points from the robot base towards the grabbing point to define grabbing orientation. Since we consider the coordinate system of the robot base as the global coordinate system, this vector corresponds to the grabbing point $\mathbf{P_G}$.

The final orientation of the end effector can be achieved on the basis of the $x$ vector $\mathbf{v_x}$, the known $y$ vector $\mathbf{v_{pl}}$ and the $z$ vector $\mathbf{v_z}$  (see Figure \ref{fig:aerial_grabbing}):
\begin{equation}
    \mathbf{R_{grabbing}}=
    \begin{bmatrix}
      \mathbf{v_x} & \mathbf{v_{pl}} & \mathbf{v_z}
    \end{bmatrix}
\end{equation}
where $v_x$ can be calculated by:
\begin{equation}
    \mathbf{v_x}=\frac{\mathbf{v_{pl}} \times \mathbf{P_G}}{\vert \mathbf{v_{pl}} \times \mathbf{P_G} \vert}
\end{equation}
and $\mathbf{v_z}$ can be calculated by:
\begin{equation}
    \mathbf{v_z}=\frac{\mathbf{v_x} \times \mathbf{v_{pl}}}{\vert \mathbf{v_x} \times \mathbf{v_{pl}} \vert}
\end{equation}

\subsection{Resolving the case for parallel magnetic field vectors measured by magnetometers}

If the magnetic field vectors of magnetometers are parallel, it is not possible to uniquely determine the conductor pose. Since the magnetometers are mounted on the  end effector of the robot, rotating the end effector by 90 degrees will change orientation of the magnetometers, resulting in non-parallel field vector measurements. In practical terms, the lines are considered parallel if the angle between them is less than $\alpha_{min}$ degrees.

\subsection{Approaching the grabbing point}

The robot cannot directly approach the conductor from any pose, and for this reason we send it first to the location $\mathbf{P_{intermittent}}$ $d$ centimeters from the final pose in the opposite direction of the vector $\mathbf{v_z}$
\begin{equation}
    \mathbf{P_{intermittent}}=\mathbf{P_G}-\mathbf{v_z}d
    \label{eq:intermittent}
\end{equation}
The orientation of this intermittent point needs to be identical to the orientation of the grabbing point $\mathbf{R_{grabbing}}$. Magnetometers on the end-effector are setup in such a way that in this pose magnetic field based localization gives the result as precise as possible. From there the robot can proceed to the grabbing point using linear movement. 

\begin{figure*}[t!]
	\centering
	\includegraphics[width=0.92\linewidth]{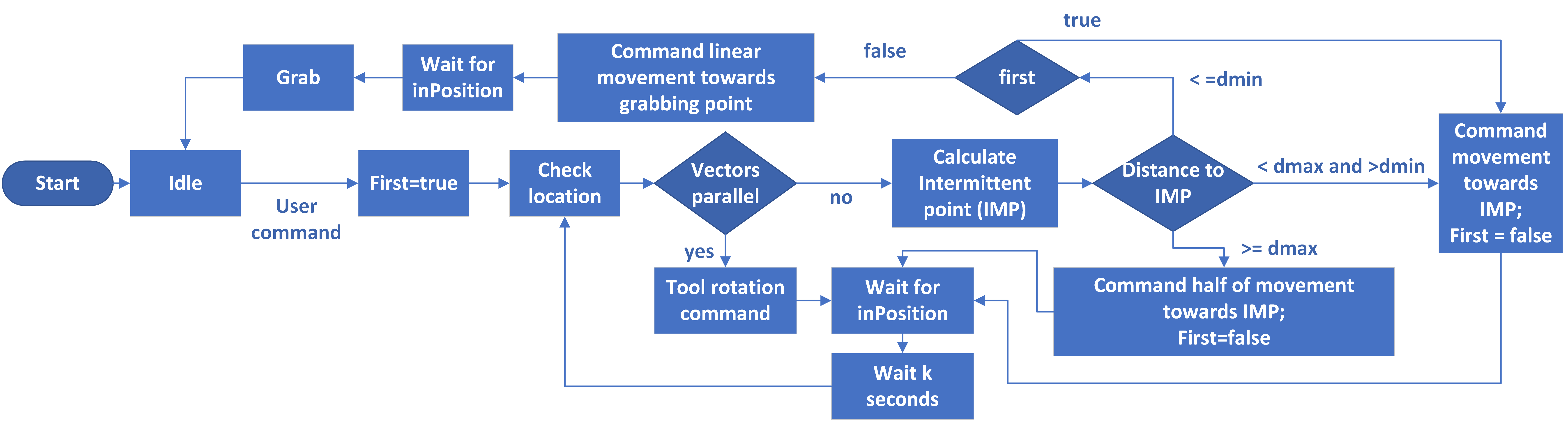}
	 \caption{Procedure for grabbing the conductor}
     \label{fig:algorithm}
\end{figure*}

To achieve autonomous grabbing of the conductor, we have created a procedure that deals with the rotation of the tool and the approach to the conductor (see Figure \ref{fig:algorithm}). 
The robot starts in idle mode. At the operator's command, the grabbing process starts where it calculates the next movement based on the known position of the conductor. If the magnetometer baseline and the conductor vector are parallel, the robot starts to rotate the end effector, waits for the robot to arrive at the commanded location, and then waits $k$ seconds to collect new information for localization. The system returns to the localization analysis. If the magnetometer baseline and the conductor vector are not parallel, an intermittent point is calculated based on \eqref{eq:intermittent}. Normally, the robot starts moving towards the intermittent point. If the distance to the intermittent point is greater than a certain distance $d_{max}$, the robot starts to move towards the midpoint between the current point and the intermittent point. In both cases, the system waits for the robot to reach its position, then waits $k$ seconds again and starts analyzing the new position. If the current distance to the intermittent point is less than a certain distance $d_{min}$, the robot starts moving linearly towards the grabbing point. When the grabbing point is reached, the conductor is grabbed and the system returns to idle mode.

\section{EXPERIMENTAL RESULTS}

We conducted experiments under laboratory conditions to verify the method and determine its precision and usability.

\subsection{Experimental setup}

\begin{figure}[th]
	\centering
	\includegraphics[width=0.54\linewidth]{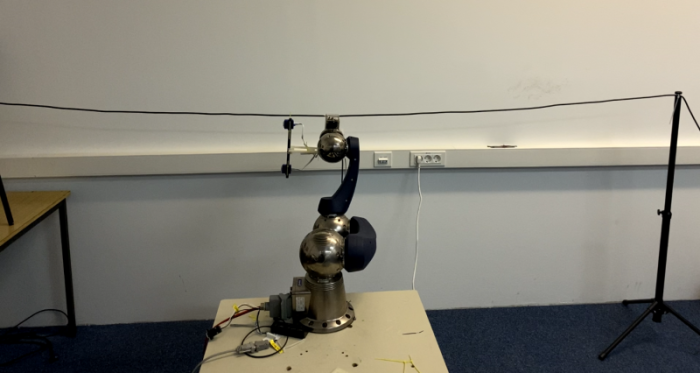}
	\includegraphics[width=0.32\linewidth]{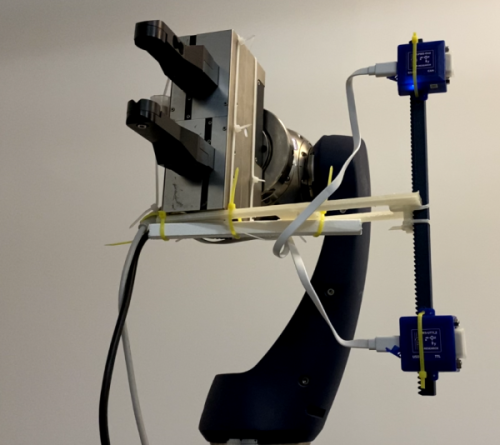}
	 \caption{Experimental setup}
     \label{fig:setup}
\end{figure}

During the experiments we used the current source with 36A effective current at 50Hz. We made a setup with 3.3m of approximately straight wire (see Figure \ref{fig:setup}). In the middle we have setup a Schunk LWA 4P, a 6-axis robot arm, equipped with two inertial measurements units (IMU), which are mounted about 15 cm from the tool center (see Figure \ref{fig:setup}) and perpendicular to the grabbing direction. The distance between the IMUs is 20 cm, and both have the same orientation (matrix $\mathbf{R}$ in \eqref{eq::magnetometertransformation} is an identity matrix). The IMUs used are LPMS-CU2 and LPMS-UTTL2, both equipped with a 3-axis magnetometer with 16 Gauss range and 200 Hz sampling rate.

During the experiments, several phenomena affect the accuracy of the proposed algorithm: the power line has a finite length and is not perfectly straight, the return conductor (distanced 2m from the setup) has some influence on the magnetometer readings, ferromagnetic materials in the robot arm and end-effector could affect the magnetic field generated by the wire.

\subsection{Localization experiments} \label{subsec:localizationexperiments}

In order to demonstrate the developed localization method based on magnetic field strength, we conducted experiments on the movement of the robot in its workspace near the conductor and recorded the results.

This experiment was performed by manually sending the robot in 13 poses one after the other (12 unique poses, since 2 of them are identical). At each point, the robot stayed for about 10s to collect data and to determine the pose of the conductor relative to the sensors and consequently to the robot base coordinate system.
The Table \ref{tab:points_for_localization} shows the individual joints for each of the 13 robot poses.

\begin{table}[]
    \centering
        \caption{Individual joints for each pose for localization}
    \begin{tabular}{|c|c|c|c|c|c|c|}
        \hline
         Pose & $q_0[\deg]$ & $q_1[\deg]$ & $q_2[\deg]$ & $q_3[\deg]$ & $q_4[\deg]$ & $q_5[\deg]$ \\
         \hline
1 &0 &-67.12 &-58.45 &0 &81.33 &90\\
\hline
2 &0 &-84.31 &-75.64 &0 &81.33 &90\\
\hline
3 &0 &-101.5 &-92.83 &0 &81.33 &90\\
\hline
4 &0 &-84.31 &-75.64 &17.19 &81.33 &90\\
\hline
5 &0 &-84.31 &-75.64 &-17.19 &81.33 &90\\
\hline
6 &0 &-84.31 &-75.64 &0 &81.33 &45\\
\hline
7 &0 &-84.31 &-75.64 &0 &81.33 &133\\
\hline
8 &0 &-67.12 &-58.45 &0 &81.33 &90\\
\hline
9 &-22.91 &-67.12 &-58.45 &0 &81.33 &90\\
\hline
10 &-45.84 &-67.12 &-58.45 &0 &81.33 &90\\
\hline
11 &-68.77 &-67.12 &-58.45 &0 &81.33 &90\\
\hline
12 &22.93 &-67.12 &-58.45 &0 &81.33 &90\\
\hline
13 &45.86 &-67.12 &-58.45 &0 &81.33 &90\\
\hline       
    \end{tabular}
    \label{tab:points_for_localization}
\end{table}

\begin{figure}[th]
	\centering
	\includegraphics[width=0.85\linewidth]{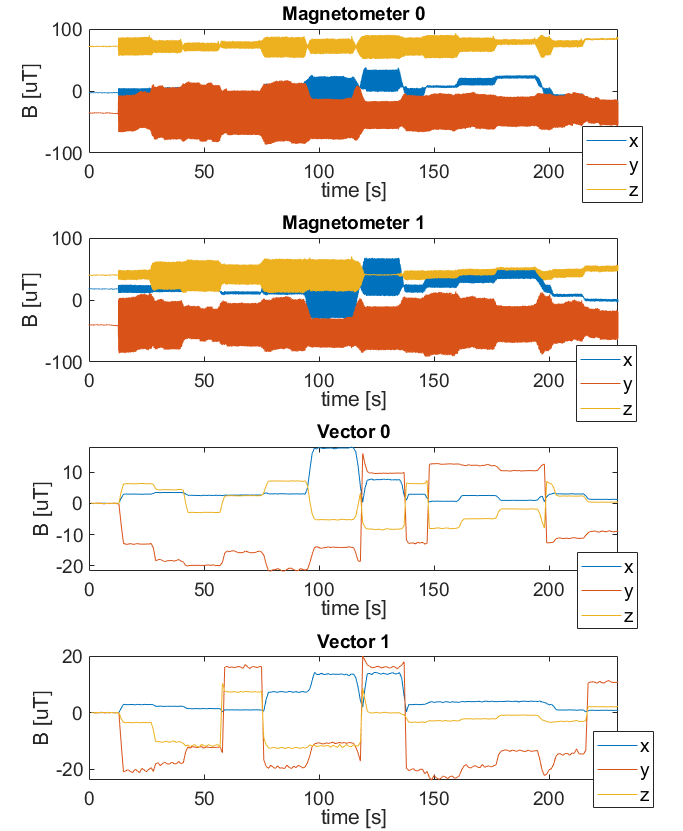}
	 \caption{Measurements of individual magnetometers and  magnetic field vectors of 50Hz signal components}
     \label{fig:vectors}
\end{figure}

The raw data measured by magnetometers is shown in first two graphs of Figure \ref{fig:vectors}. By processing this data with the methods shown in section \ref{sec:localization}, magnetic field vectors of 50 Hz signal components can be determined. During the experiments, we used 200 samples for Fourier transform analysis, and since the sampling frequency is 200 Hz, the desired vector is calculated every second. The vectors can be seen in the figure \ref{fig:vectors}.



\begin{figure}[th]
	\centering
	\includegraphics[width=0.445\linewidth]{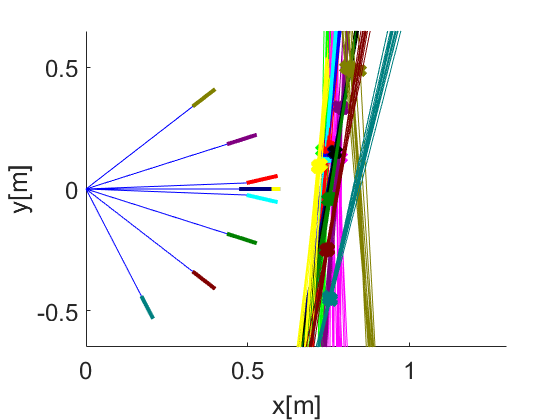}
	\includegraphics[width=0.445\linewidth]{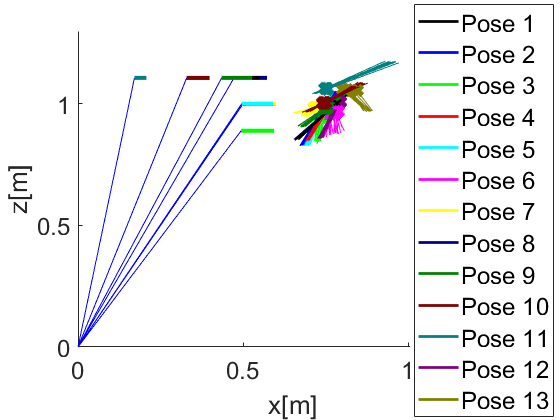}
	\includegraphics[width=0.445\linewidth]{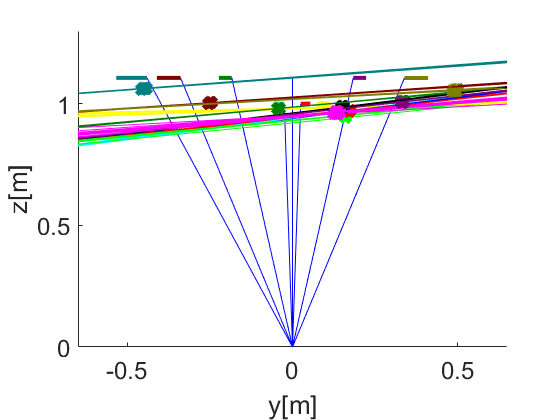}
	\includegraphics[width=0.445\linewidth]{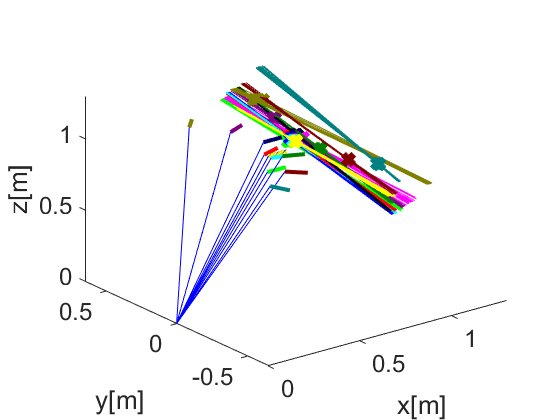}
	 \caption{Estimated locations of conductor from different poses. Estimations of the conductor locations for each robot pose are shown as lines with different colors.}
     \label{fig:conductor_location}
\end{figure}

Figure \ref{fig:conductor_location} shows the estimated poses of the conductor based on the magnetic field strength together with the positions where the robot end-effector was located for each pose. The estimated positions during the transition between two poses are not shown because they do not provide useful information. For each pose, we collected several estimated positions of the conductor, and they are all colored in the same color. Table \ref{tab:vectors} shows the average value for each pose and its standard deviation.
\begin{table}[]
    \centering
    \caption{The average and the standard deviation of the conductor vector}
    \begin{tabular}{|c|c|c|c|c|c|c|c|}
    \hline
    Pose & Cnt & $\overline{x}$ & $\overline{y}$& $\overline{z}$ & $\sigma(x)$ & $\sigma(y)$ & $\sigma(z)$ \\
    \hline
1 &12 &-0.131 &-0.979 &-0.159 &0.005 &0.000 &0.002 \\
\hline
2 &11 &-0.078 &-0.983 &-0.168 &0.003 &0.000 &0.001 \\
\hline
3 &14 &-0.016 &-0.991 &-0.130 &0.002 &0.000 &0.002 \\
\hline
4 &15 &-0.063 &-0.987 &-0.148 &0.004 &0.000 &0.001 \\
\hline
5 &15 &-0.059 &-0.984 &-0.169 &0.005 &0.000 &0.002 \\
\hline
6 &17 &0.007 &-0.994 &-0.109 &0.015 &0.001 &0.011 \\
\hline
7 &12 &-0.064 &-0.997 &-0.038 &0.016 &0.001 &0.008 \\
\hline
8 &6 &-0.131 &-0.978 &-0.161 &0.004 &0.000 &0.002 \\
\hline
9 &12 &-0.120 &-0.985 &-0.122 &0.006 &0.001 &0.002 \\
\hline
10 &13 &-0.128 &-0.988 &-0.090 &0.009 &0.001 &0.002 \\
\hline
11 &14 &-0.175 &-0.980 &-0.097 &0.009 &0.002 &0.002 \\
\hline
12 &11 &-0.055 &-0.987 &-0.149 &0.005 &0.000 &0.001 \\
\hline
13 &12 &0.053 &-0.996 &-0.075 &0.009 &0.001 &0.001 \\
\hline
\hline
All &164 &-0.070 &-0.987 &-0.123 &0.063 &0.006 &0.038 \\

\hline
\end{tabular}
    \label{tab:vectors}
\end{table}

\subsection{Grabbing experiments}

In order to demonstrate the usability of the localization system shown in \ref{subsec:localizationexperiments}, we performed grabbing experiments. In these experiments, we did not operate the gripper, since the goal was to have a gripper envelope the conductor. We have conducted grabbing experiments for all 12 individual robot starting poses shown in table \ref{tab:points_for_localization} (table has 13 poses, but since pose 1 is identical to pose 8, it was skipped).

\begin{figure}[th]
	\centering
	\includegraphics[width=0.445\linewidth]{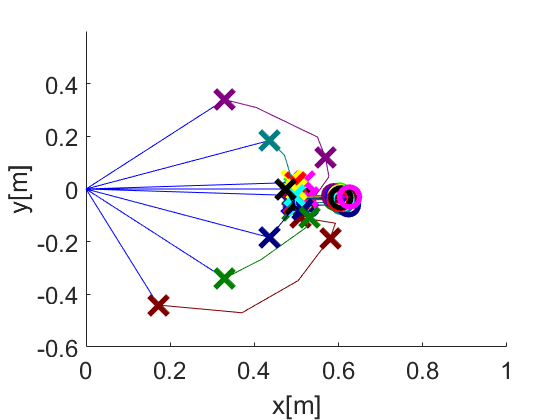}
	\includegraphics[width=0.445\linewidth]{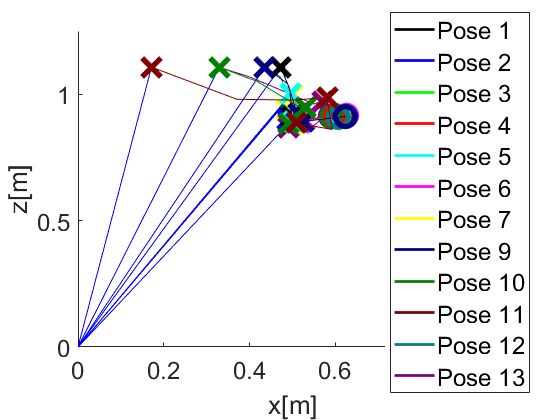}
	\includegraphics[width=0.445\linewidth]{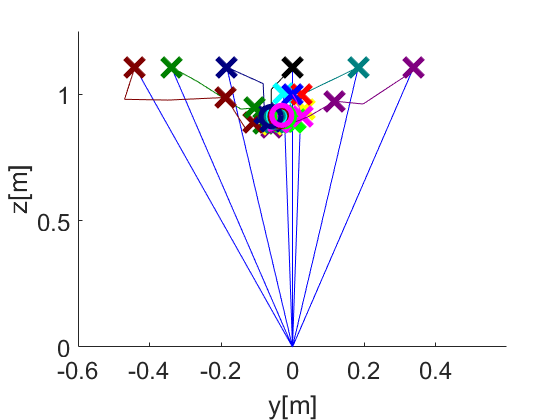}
	\includegraphics[width=0.445\linewidth]{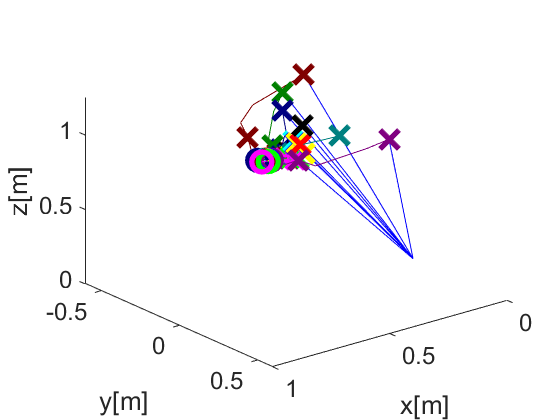}
	 \caption{Movement of the end-effector during grabbing the conductor. $\mathbf{x}$ symbols stopping point, while $\mathbf{o}$ symbols the final point for each pose}
     \label{fig:gripping}
\end{figure}

\begin{figure}[th]
	\centering
	\includegraphics[width=0.21\linewidth]{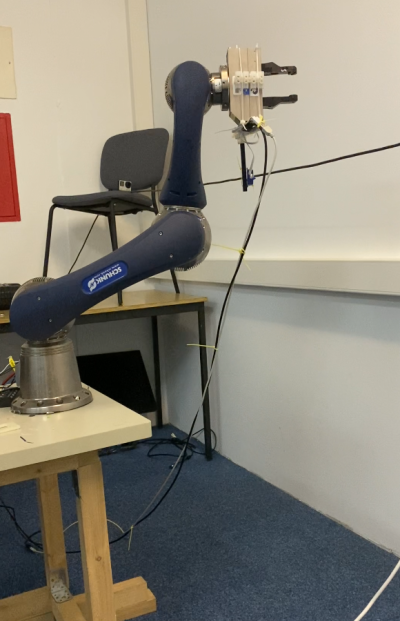}
	\includegraphics[width=0.21\linewidth]{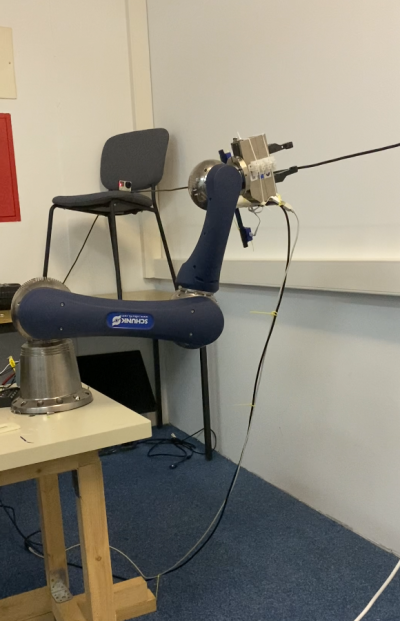}
	\includegraphics[width=0.21\linewidth]{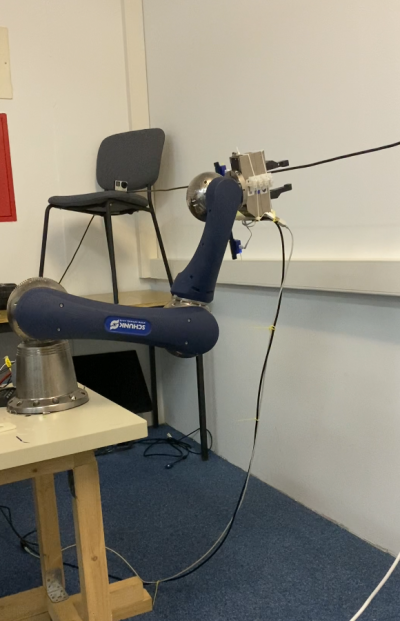}
	\includegraphics[width=0.21\linewidth]{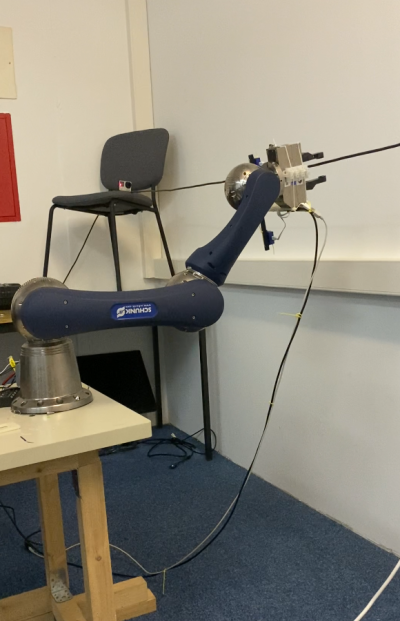}
	 \caption{Example of robot movement during grabbing procedure}
     \label{fig:robot}
\end{figure}

\begin{figure}[H]
	\centering
	\includegraphics[width=0.46\linewidth]{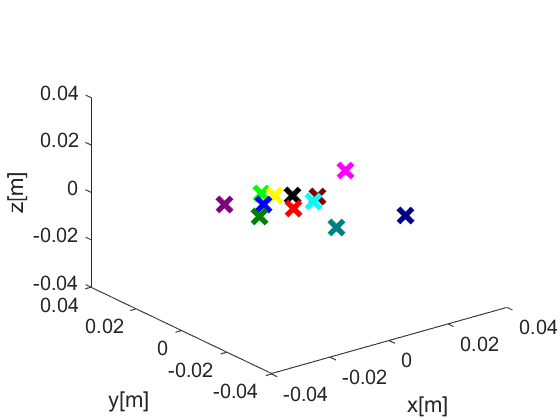}
	\includegraphics[width=0.46\linewidth]{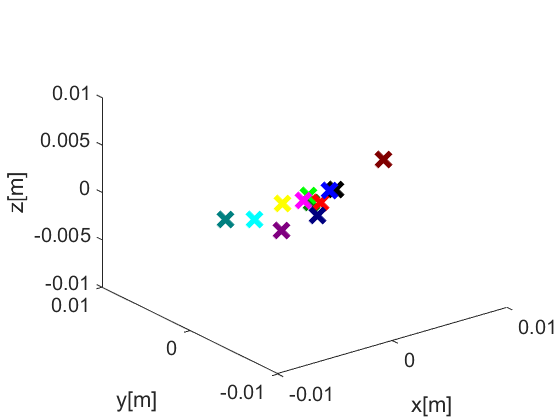}
	 \caption{Error for each final pose, compared to the average of all final poses. Left figure shows errors for 12 different starting poses and right figure shows error for 12 experiment with the same starting pose}
     \label{fig:error}
\end{figure}

The movement of the end-effector for each of the experiments is shown in the figure \ref{fig:gripping}
Since for these experiments, the conductor is located at the same position, and robot base doesn't move, the grabbing point, as well as, last intermittent point, should both approximately correspond regardless of the starting pose. Left part of figure \ref{fig:error}  shows the error for each final grabbing pose, compared to the average of all final poses. Table \ref{tab:gripping} shows the number of stopping points during grabbing and error of the individual final pose compared to the average of all final poses. The example of robot movement during experiment is shown in Figure \ref{fig:robot}. The results have shown that the final deviation for each position is less than 3cm which is enough for grabbing the conductor.

To test the repeatability of the system, we have tested the grabbing procedure 12 times from start position 4 in Table \ref{tab:points_for_localization}. The error of each end pose from the average of all end poses is shown on the right side of the Figure \ref{fig:error}. The maximum error is less than 1 cm and the standard deviation is 3.4 mm.


\begin{table}[]
    \caption{The number of stopping points (SP) during grabbing and error of the individual final pose compared to the average of all final poses.}

    \centering
    \begin{tabular}{|c|c|c|c|c|c|c|c|c|c|c|c|c|}
    \hline
Pose &1 &2 &3 &4 &5 &6 &7 &8 &9 &10 &12 &13 \\
\hline
SP &3  &3  &3  &3  &3  &4  &4  &3  &4  &5  &3  &4  \\
\hline
Err &3  &10  &14  &6  &4  &22  &8  &30  &14  &8  &16  &20  \\
(mm) &  &  &  &  &  &  &  &  &  &  &  &  \\
\hline
    \end{tabular}
    \label{tab:gripping}
\end{table}

\section{CONCLUSIONS}

In this paper we have presented a method for grabbing straight conductor carrying alternating current based only on the magnetic field strength. We have shown a conductor localization method based on two magnetometers and the method for grabbing. The method was demonstrated experimentally under laboratory conditions with satisfactory results.

The next steps will include an attempt to speed up the signal processing of AC by using a smaller number of samples required to compute the AC signal amplitude to achieve a position update rate of at least 20 times per second, allowing continuous movement towards the grabbing point. We also plan to experiment with two parallel conductors and try to grab one of them. Finally, the goal is to realise the grabbing of the power line conductor using aerial manipulation.

\addtolength{\textheight}{-12cm}   







\bibliographystyle{IEEEtran}

\bibliography{bibliography.bib}

\begin{thebibliography}{10}
\providecommand{\url}[1]{#1}
\csname url@rmstyle\endcsname
\providecommand{\newblock}{\relax}
\providecommand{\bibinfo}[2]{#2}
\providecommand\BIBentrySTDinterwordspacing{\spaceskip=0pt\relax}
\providecommand\BIBentryALTinterwordstretchfactor{4}
\providecommand\BIBentryALTinterwordspacing{\spaceskip=\fontdimen2\font plus
\BIBentryALTinterwordstretchfactor\fontdimen3\font minus \fontdimen4\font\relax}
\providecommand\BIBforeignlanguage[2]{{%
\expandafter\ifx\csname l@#1\endcsname\relax
\typeout{** WARNING: IEEEtran.bst: No hyphenation pattern has been}%
\typeout{** loaded for the language `#1'. Using the pattern for}%
\typeout{** the default language instead.}%
\else
\language=\csname l@#1\endcsname
\fi
#2}}

\bibitem{ollero2018aeroarms}
A.~Ollero, G.~Heredia, A.~Franchi, G.~Antonelli, K.~Kondak, A.~Sanfeliu, A.~Viguria, J.~R. Martinez-de Dios, F.~Pierri, J.~Cort{\'e}s, \emph{et~al.}, ``The aeroarms project: Aerial robots with advanced manipulation capabilities for inspection and maintenance,'' \emph{IEEE Robotics \& Automation Magazine}, vol.~25, no.~4, pp. 12--23, 2018.

\bibitem{korpela2014hardware}
C.~Korpela, M.~Orsag, and P.~Oh, ``Hardware-in-the-loop verification for mobile manipulating unmanned aerial vehicles,'' \emph{Journal of intelligent \& robotic systems}, vol.~73, no. 1-4, pp. 725--736, 2014.

\bibitem{brown1973}
M.~Brown Boveri \& Cie~AG, ``Verfarhen und vorrichtung zur selbsta\"{a}tigen spurf\"{u}hrung von gleiselosen fahrzeugen,'' \emph{German pattent no: 2137631}, 1973.

\bibitem{Everett1995}
H.~Everett, \emph{Sensors for mobile robot: Theory and application}.\hskip 1em plus 0.5em minus 0.4em\relax Wellesley, Massachusetts, NJ, USA: A K Peters, Ltd., 1995.

\bibitem{Kamewaka1987}
S.~Kamewaka and S.~Uemura, ``A magnetic guidance method for automated guided vehicles,'' \emph{IEEE Transactions on Magnetics}, vol.~23, no.~5, pp. 2416--2418, 1987.

\bibitem{martinovic_electric_2014}
D.~Martinovic, M.~Grimm, and H.-C. Reuss, ``\BIBforeignlanguage{English}{Electric {Vehicle} {Positioning} {Concept} for {Inductive} {Charging} {Purposes} {Using} {Magnetic} {Fields}},'' in \emph{\BIBforeignlanguage{English}{{IEEE} {Power} \& {Energy} {Student} {Summit} 2014}}.\hskip 1em plus 0.5em minus 0.4em\relax Ostfildern: haka print und medien GmbH, 2014, pp. 11--16.

\bibitem{martinovic_magnetic_2015}
D.~Martinovic, C.~Binz, and H.-C. Reuss, ``\BIBforeignlanguage{English}{Magnetic {Field} based {Localization} of the {Charging} {Coil} using {Trilateration}},'' in \emph{\BIBforeignlanguage{English}{Autoreg 2015 - {VDI}-{Berichte} 2233}}.\hskip 1em plus 0.5em minus 0.4em\relax Düsseldorf: VDI Verlag GmbH, 2015, pp. 129--140.

\bibitem{martinovic_electric_2014-1}
D.~Martinovic, M.~Grimm, and H.-C. Reuss, ``\BIBforeignlanguage{English}{Electric {Vehicle} {Positioning} for {Inductive} {Charging} {Purposes} {Using} {Magnetic} {Field} {Distortion} {Elimination} in {High}-{Permeability} {Environments}},'' \emph{\BIBforeignlanguage{English}{IEEE Transactions on Magnetics}}, vol.~50, no.~11, pp. 1--4, Nov. 2014.

\bibitem{martinovic_dealing_2019}
D.~Martinovic, ``Dealing with magnetization effects in {EV} positioning systems based on periodic magnetic signals,'' \emph{AIP Advances}, vol.~9, no.~3, p. 035214, Mar. 2019, publisher: American Institute of Physics.

\bibitem{Sheinker2013}
A.~Sheinker, B.~Ginzburg, N.~Salomonski, L.~Frumkis, and B.~Z. Kaplan, ``{Localization in 2D using beacons of low frequency magnetic field},'' \emph{IEEE Journal of Selected Topics in Applied Earth Observations and Remote Sensing}, vol.~6, no.~2, pp. 1020--1030, 2013.

\bibitem{Sheinker20132}
------, ``{Localization in 3-D using beacons of low frequency magnetic field},'' \emph{IEEE Transactions on Instrumentation and Measurement}, vol.~62, no.~12, pp. 3194--3201, 2013.

\bibitem{Sheinker2016}
A.~Sheinker, B.~Ginzburg, N.~Salomonski, L.~Frumkis, B.~Z. Kaplan, and M.~B. Moldwin, ``{A method for indoor navigation based on magnetic beacons using smartphones and tablets},'' \emph{Measurement: Journal of the International Measurement Confederation}, vol.~81, pp. 197--209, 2016.

\bibitem{Mitterer2018}
T.~Mitterer, H.~Gietler, L.-M. Faller, and H.~Zangl, ``{Artificial Landmarks for Autonomous Vehicles Based on Magnetic Sensors},'' \emph{Proceedings}, vol.~2, no.~13, p. 856, 2018.

\bibitem{Son2016}
D.~Son, S.~Yim, and M.~Sitti, ``{A 5-D Localization Method for a Magnetically Manipulated Untethered Robot Using a 2-D Array of Hall-Effect Sensors},'' \emph{IEEE/ASME Transactions on Mechatronics}, vol.~21, no.~2, pp. 708--716, 2016.

\bibitem{Than2012}
T.~D. Than, G.~Alici, H.~Zhou, and W.~Li, ``{A review of localization systems for robotic endoscopic capsules},'' \emph{IEEE Transactions on Biomedical Engineering}, vol.~59, no.~9, pp. 2387--2399, 2012.

\bibitem{Song2014}
S.~Song, B.~Li, W.~Qiao, C.~Hu, H.~Ren, H.~Yu, Q.~Zhang, M.~Q. Meng, and G.~Xu, ``{6-D magnetic localization and orientation method for an annular magnet based on a closed-form analytical model},'' \emph{IEEE Transactions on Magnetics}, vol.~50, no.~9, 2014.

\bibitem{Popek2017}
K.~M. Popek, T.~Schmid, and J.~J. Abbott, ``{Six-Degree-of-Freedom Localization of an Untethered Magnetic Capsule Using a Single Rotating Magnetic Dipole},'' \emph{IEEE Robotics and Automation Letters}, vol.~2, no.~1, pp. 305--312, 2017.

\bibitem{Hu2005}
C.~Hu, M.~Q. Meng, and M.~Mandal, ``{Efficient magnetic localization and orientation technique for capsule endoscopy},'' \emph{2005 IEEE/RSJ International Conference on Intelligent Robots and Systems, IROS}, pp. 628--633, 2005.

\bibitem{Hu2007}
------, ``{A linear algorithm for tracing magnet position and orientation by using three-axis magnetic sensors},'' \emph{IEEE Transactions on Magnetics}, vol.~43, no.~12, pp. 4096--4101, 2007.

\bibitem{hu2010}
C.~Hu, M.~Li, S.~Song, W.~Yang, R.~Zhang, and M.~Q. Meng, ``{A Cubic 3-Axis Magnetic Sensor Array for Wirelessly Tracking Magnet Position and Orientation},'' \emph{IEEE Sensors Journal}, vol.~10, no.~5, pp. 903--913, 2010.

\bibitem{Vallivaara2011}
I.~Vallivaara, J.~Haverinen, A.~Kemppainen, and J.~R{\"{o}}ning, ``{Magnetic field-based SLAM method for solving the localization problem in mobile robot floor-cleaning task},'' \emph{IEEE 15th International Conference on Advanced Robotics: New Boundaries for Robotics, ICAR 2011}, pp. 198--203, 2011.

\bibitem{Wang2016}
S.~Wang, H.~Wen, R.~Clark, and N.~Trigoni, ``{Keyframe based large-scale indoor localisation using geomagnetic field and motion pattern},'' \emph{IEEE International Conference on Intelligent Robots and Systems}, vol. 2016-November, pp. 1910--1917, 2016.

\bibitem{Lee2018}
N.~Lee, S.~Ahn, and D.~Han, ``{AMID: Accurate magnetic indoor localization using deep learning},'' \emph{Sensors (Switzerland)}, vol.~18, no.~5, 2018.

\bibitem{Akai2017}
N.~Akai and K.~Ozaki, ``{3D magnetic field mapping in large-scale indoor environment using measurement robot and Gaussian processes},'' \emph{2017 International Conference on Indoor Positioning and Indoor Navigation, IPIN 2017}, vol. 2017-Janua, no. September, pp. 1--7, 2017.

\bibitem{Frassl2013}
M.~Frassl, M.~Angermann, M.~Lichtenstern, P.~Robertson, B.~J. Julian, and M.~Doniec, ``{Magnetic maps of indoor environments for precise localization of legged and non-legged locomotion},'' \emph{IEEE International Conference on Intelligent Robots and Systems}, no. May 2014, pp. 913--920, 2013.

\bibitem{Hanley2017}
D.~Hanley, A.~B. Faustino, S.~D. Zelman, D.~A. Degenhardt, and T.~Bretl, ``{MagPIE: A dataset for indoor positioning with magnetic anomalies},'' \emph{2017 International Conference on Indoor Positioning and Indoor Navigation, IPIN 2017}, vol. 2017-Janua, pp. 1--8, 2017.

\bibitem{martinovic_mathematical_2021}
D.~Martinović, S.~Bogdan, and Z.~Kovačić, ``\BIBforeignlanguage{en}{Mathematical {Considerations} for {Unmanned} {Aerial} {Vehicle} {Navigation} in the {Magnetic} {Field} of {Two} {Parallel} {Transmission} {Lines}},'' \emph{\BIBforeignlanguage{en}{Applied Sciences}}, vol.~11, no.~8, p. 3323, Jan. 2021, number: 8.

\bibitem{Richtsfeld2008}
M.~Richtsfeld, W.~Ponweiser, and M.~Vincze, ``{Real time grasping of freely placed cylindrical objects},'' \emph{ICINCO 2008 - Proceedings of the 5th International Conference on Informatics in Control, Automation and Robotics}, vol. 1 RA, pp. 165--170, 2008.

\bibitem{Sintov2019}
A.~Sintov, O.~Tslil, and S.~Frenkel, ``{Longitudinal Regrasping of Elongated Objects},'' \emph{Robotica}, vol.~38, pp. 639--651, 2019.

\bibitem{Koo2011}
K.~Koo, X.~Jiang, A.~Konno, and M.~Uchiyama, ``{Development of a wire harness assembly motion planner for redundant multiple manipulators},'' \emph{Journal of Robotics and Mechatronics}, vol.~23, no.~6, pp. 907--918, 2011.

\bibitem{Qin2019}
Y.~Qin, A.~Escande, and E.~Yoshida, ``{Cable Installation by a Humanoid Integrating Dual-arm Manipulation and Walking},'' \emph{Proceedings of the 2019 IEEE/SICE International Symposium on System Integration, SII 2019}, pp. 98--103, 2019.

\bibitem{Zapata-Impata2019}
B.~S. Zapata-Impata, P.~Gil, J.~Pomares, and F.~Torres, ``{Fast geometry-based computation of grasping points on three-dimensional point clouds},'' \emph{International Journal of Advanced Robotic Systems}, vol.~16, no.~1, pp. 1--18, 2019.

\bibitem{Seo2017}
H.~Seo, S.~Kim, and H.~J. Kim, ``{Aerial grasping of cylindrical object using visual servoing based on stochastic model predictive control},'' \emph{Proceedings - IEEE International Conference on Robotics and Automation}, pp. 6362--6368, 2017.

\end{thebibliography}

\end{document}